\documentclass[12pt,letterpaper]{article}

\usepackage[margin=1in]{geometry}
\usepackage{amsmath,amssymb,amsthm}
\usepackage{booktabs}
\usepackage{array}
\usepackage{graphicx}
\usepackage{xcolor}
\usepackage{microtype}
\usepackage[numbers,super,sort&compress]{natbib}
\usepackage[pdfencoding=auto,psdextra,colorlinks=true,citecolor=blue,linkcolor=blue,urlcolor=blue]{hyperref}
\usepackage{lineno}
\usepackage{setspace}
\usepackage{enumitem}
\usepackage{caption}
\usepackage{placeins}

\theoremstyle{plain}

\newtheorem{proposition}{Proposition}

\graphicspath{{figures/}}

\newcommand{\kstar}{k_{\star}}
\newcommand{\PB}{P_B}
\newcommand{\rstar}{r_{\star}}

\newcolumntype{L}[1]{>{\raggedright\arraybackslash}p{#1}}
\DeclareMathOperator{\spanop}{span}
\DeclareMathOperator{\Cert}{Cert}
\DeclareMathOperator{\Amb}{Amb}
\DeclareMathOperator{\cost}{cost}

\hypersetup{
  pdftitle={Deployment-complete benchmarking},
  pdfauthor={El Mustapha Mansouri and Keigo Arai},
  pdfsubject={A framework for auditing whether benchmark evidence determines deployment actions, using claim-completeness, mixed-fiber audits and completion curves across machine learning, toxicology and materials discovery.},
  pdfkeywords={deployment-complete benchmarking, machine learning benchmarking, benchmark completeness, benchmark evidence, deployment actions, benchmark-to-deployment gap, evidence fibers, mixed fibers, completion curves, completion cost, response-rank certification, conformal prediction, calibration, uncertainty quantification, out-of-distribution detection, active learning, value of information, selective classification, trustworthy machine learning, model evaluation, deployment risk, scientific screening, materials discovery, toxicity prediction, Tox21, JARVIS, Matbench Discovery},
  pdflang={en-US},
  pdfcreator={LaTeX},
}

\begin{document}
\hypersetup{pageanchor=false}

\begin{titlepage}
\vspace*{2cm}
\begin{center}
{\LARGE\bfseries Deployment-complete benchmarking}

\vspace{1.3cm}
{\large El~Mustapha~Mansouri$^{1,*}$,\enspace Keigo~Arai$^{1}$}

\vspace{0.7cm}
{\normalsize
$^{1}$School of Engineering, Institute of Science Tokyo,
Yokohama, Kanagawa 226-8501, Japan\\[6pt]
$^{*}$Correspondence: \texttt{mansouri.e.2224@m.isct.ac.jp}}
\end{center}
\vfill
\end{titlepage}

\newpage

\begin{abstract}
\thispagestyle{empty}
\noindent
Benchmarks increasingly guide deployment, procurement and scientific screening,
yet a score supports only the response it records, not necessarily the
deployment action. We introduce deployment-complete benchmarking, which tests
whether benchmark evidence determines a deployment action. A benchmark is
complete for a claim exactly when the action is constant on each evidence
fiber; mixed fibers expose missing deployment information, and completion
curves quantify the evidence required to resolve ambiguity. In controlled
response spaces, benchmark-channel conformal coverage of 94.98\% transferred
poorly to an unmeasured deployment channel (10.07\%), whereas response-rank
intervals achieved 94.91\% coverage; even zero benchmark error certified only
45.4\% of candidates at the largest residual size. Public audits revealed
incompleteness, including 97.9\% mixed Tox21 fibers and zero median
certifiable fraction in main Matbench and JARVIS audits. In held-out replays,
certify-then-acquire reduced false decisions from 1.19\% to 0.027\% in Tox21
and from 20.3\% to 0.128\% in JARVIS, while changing model choice and
identifying deployment-relevant probes. Deployment-ready benchmarks should
report evidence, supported actions, ambiguity and completion cost rather than
scores alone.
\end{abstract}

\newpage
\hypersetup{pageanchor=true}
\pagenumbering{arabic}
\setcounter{page}{1}

Benchmarks rank models, guide procurement and increasingly determine which
systems are trusted outside the test set. A score therefore often becomes
evidence for an action: deploy a model, advance a material, triage a compound
or decide which expensive measurement to acquire next.

But a benchmark score is evidence only for the response it records. Deployment
may depend on another response, such as robustness, safety, toxicity,
stability or a thresholded physical property. A model can be accurate,
calibrated and competitive on the benchmark while the action made from that
score remains underdetermined.

We introduce deployment-complete benchmarking. A benchmark records evidence
$E:\mathcal{S}\to\mathcal{Z}$, and a deployment claim is an action
$D:\mathcal{S}\to\mathcal{A}$. Here $\mathcal{S}$, $\mathcal{Z}$ and
$\mathcal{A}$ denote candidates, evidence and actions. The benchmark is
complete for the claim when candidates with the same evidence require the same
action, equivalently $D=\phi\circ E$. Mixed fibers are witnesses of missing
deployment information.

This gives a diagnostic, a limit and a design rule. The diagnostic is the
mixed-fiber audit; the limit is that no statistic computed from the same
evidence can complete a missing-response claim; the design rule is the
completion curve, which measures the cost of acquiring the information needed
to support the action. Across controlled spaces, public audits and held-out
replays, benchmark accuracy, calibration and uncertainty can leave deployment
actions unresolved; completion evidence reduces false decisions, changes model
choice and identifies the missing response to measure.

\section*{Results}

\subsection*{Benchmark evidence determines only some deployment claims}

\begin{figure}[t]
\centering
\includegraphics[width=\linewidth]{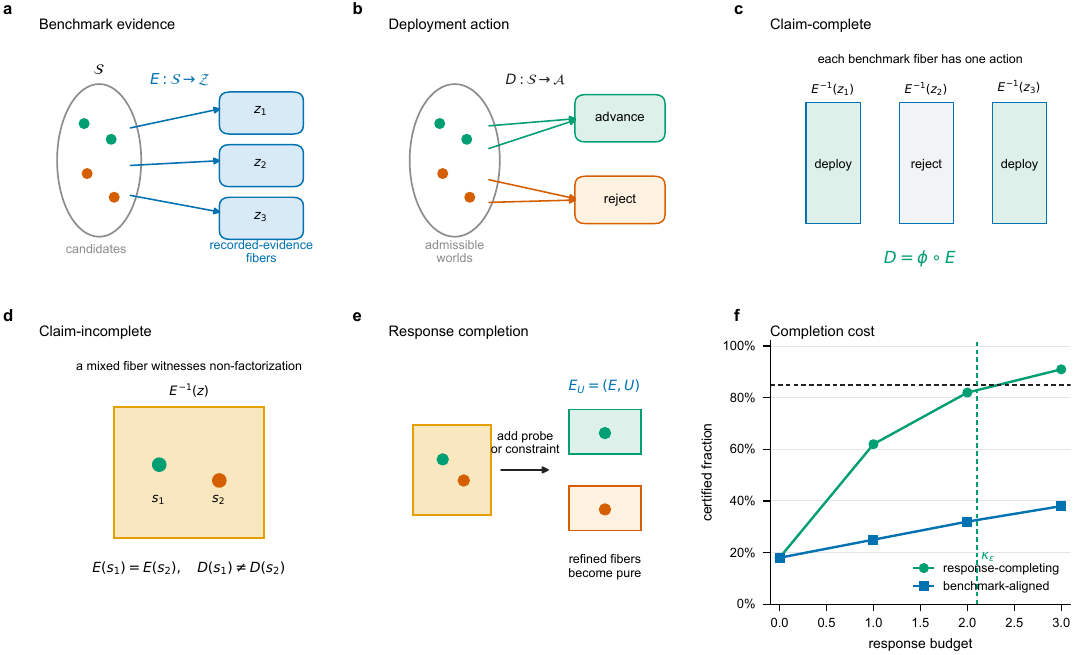}
\caption{\textbf{Deployment-complete benchmarking tests whether benchmark
evidence supports an action.} \textbf{a}, A benchmark report records an
evidence map $E:\mathcal{S}\to\mathcal{Z}$ that partitions candidates into
fibers. \textbf{b}, A deployment claim specifies an action
$D:\mathcal{S}\to\mathcal{A}$ that may depend on responses outside the score.
\textbf{c}, The benchmark is claim-complete when the action is constant on
every evidence fiber, equivalently $D=\phi\circ E$. \textbf{d}, A mixed fiber
containing benchmark twins with different deployment actions witnesses missing
evidence. \textbf{e}, Adding a response probe or trusted constraint $U$ refines
the evidence map to $E_U=(E,U)$ and can make fibers pure. \textbf{f},
Completion curves report the certified fraction as a function of response
budget; the completion cost $\kappa_\epsilon$ is the budget needed to reach a
target certified fraction.}
\label{fig:standard}
\end{figure}

Deployment-complete benchmarking turns the benchmark-to-deployment step into an
auditable property of a report (Fig.~\ref{fig:standard}). A benchmark records an evidence map
$E:\mathcal{S}\to\mathcal{Z}$, while the deployment claim is an action
$D:\mathcal{S}\to\mathcal{A}$. The benchmark is claim-complete for $D$ exactly
when the action is constant on every benchmark fiber. Equivalently,
$D=\phi\circ E$. A mixed fiber is therefore a mathematical witness that the
reported evidence package does not yet support the claim.

\begin{proposition}[Claim-completeness diagnostic]
Let $E:\mathcal{S}\to\mathcal{Z}$ be benchmark evidence and
$D:\mathcal{S}\to\mathcal{A}$ a deployment action. The following are
equivalent: $D$ is determined by benchmark evidence; $D$ is constant on every
fiber $E^{-1}(z)$; there exists a map $\phi:\mathcal{Z}\to\mathcal{A}$ such
that $D=\phi\circ E$; and no two benchmark-indistinguishable admissible worlds
require different deployment actions.
\end{proposition}

The consequence is a benchmark reporting standard. A deployment-facing
benchmark should not report a score alone; it should report the action the
score is meant to support, the evidence map used to support it, the fraction of
candidates whose action is already determined and the completion cost of the
remaining ambiguity. This turns leaderboard evaluation into an evidence-design
problem.

A statistic computed only from the same declared evidence cannot repair
non-factorization. If $T=h(E)$ and $D$ does not factor through $E$, then $D$
cannot factor through $T$, because $D=\psi\circ h\circ E$ would imply
factorization through $E$. This applies to leaderboard scores, calibration
summaries or uncertainty summaries only when they are functions of $E$ alone.
Methods using new measurements, descriptors, model internals, auxiliary labels
or structural assumptions refine the evidence map instead of repairing
non-factorization by scoring.

The resulting comparison with standard evaluation objects is simple
(Table~\ref{tab:evaluation-comparison}): they measure performance or reliability
for a recorded response, whereas deployment completeness asks what actions the
evidence package supports.

\begin{table}[t]
\centering
\small
\begingroup
\renewcommand{\arraystretch}{1.16}
\setlength{\tabcolsep}{5pt}
\begin{tabular}{L{0.24\linewidth}L{0.34\linewidth}L{0.33\linewidth}}
\toprule
Evaluation object & Question answered & What remains unresolved \\
\midrule
Accuracy or leaderboard &
Which model predicts recorded labels best? &
Whether the recorded labels determine the deployment action \\
\addlinespace[2pt]
Calibration or conformal prediction &
Are predictions reliable for the calibrated response? &
Whether another response or action is covered \\
\addlinespace[2pt]
Uncertainty or out-of-distribution (OOD) detection &
Is the input, prediction or representation unusual? &
Whether ordinary evidence fibers contain multiple actions \\
\addlinespace[2pt]
Active learning &
Which label would improve a model? &
Which response makes the deployment action decidable \\
\addlinespace[2pt]
Deployment-complete benchmarking &
Is the action a function of benchmark evidence? &
The completion cost of unresolved claims \\
\bottomrule
\end{tabular}
\endgroup
\caption{\textbf{Deployment-complete benchmarking evaluates a different
object from standard benchmark summaries.} Conventional evaluation objects
remain useful for ranking, reliability and data acquisition. Completion asks
whether the evidence package supports the action being claimed and what
response information remains missing.}
\label{tab:evaluation-comparison}
\end{table}

The design object is the completion curve: the supported fraction of deployment
claims as a function of additional evidence budget. In finite audits this is
the fraction of pure augmented-evidence fibers; in modeled response spaces it
is the corresponding declared certificate. Thus a benchmark is
deployment-ready not when its score is high, but when its completion cost is
low for the claim being made. The loss-aware and linear-response versions of
this construction are given in Methods.

\subsection*{Perfect benchmark evidence can leave deployment incomplete}

\begin{figure}[t]
\centering
\includegraphics[width=\linewidth]{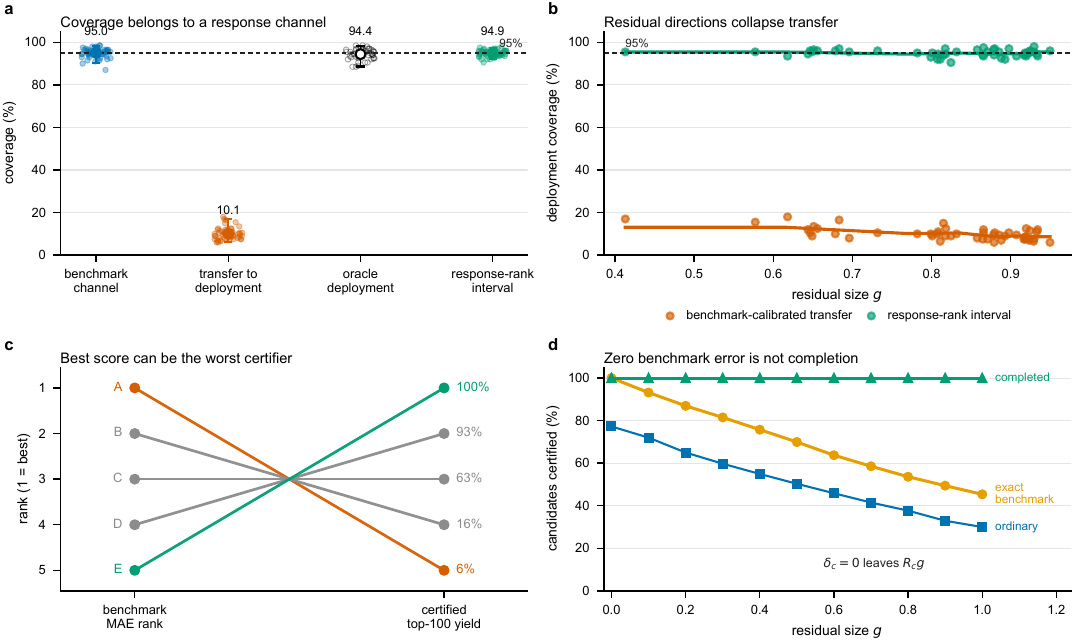}
\caption{\textbf{Perfect benchmark evidence can leave deployment incomplete.} \textbf{a},
Benchmark-channel conformal coverage does not transfer to an unmeasured
deployment response, whereas response-rank intervals recover nominal coverage
without deployment calibration labels. Points show 50 repeated
geometries/seeds; centre markers show means and error bars show 2.5--97.5
percentile ranges; dashed line marks 95\% nominal coverage. \textbf{b},
Deployment coverage under benchmark-calibrated transfer falls as the residual
response component grows, while response-rank intervals remain near 95\%
across the same 50 runs. \textbf{c}, A leaderboard inversion over 50 repeated
model-ranking runs: the best benchmark-MAE model is the worst certifier of top
deployment candidates. \textbf{d}, Zero benchmark-channel error removes
$\delta_c$ but not residual ambiguity in the controlled negative control: when
$g>0$, exact benchmark responses still leave the $R_cg$ term, whereas response
completion keeps the claim certified.}
\label{fig:main-results}
\end{figure}

Benchmark-channel conformal prediction retained 94.98\% coverage on measured
labels but only 10.07\% on the deployment channel; at the largest residual
size, exact benchmark responses still certified only 45.4\%. We isolated this
mechanism in a synthetic response space whose
geometry is known. Benchmark probes spanned three effective dimensions, while
the deployment probe varied from nearly in-span to mostly out-of-span. In this
controlled response-space experiment, $Y_\star(c)$ is the deployment response
for candidate $c$, $\hat y_c$ its benchmark-based center, $g$ the
benchmark-null residual size, $R_c$ the admissible fiber radius and $\delta_c$
the benchmark-channel error bound. Across 50 geometries and data seeds, an
oracle conformal predictor calibrated on deployment labels recovered 94.42\%,
and response-rank intervals using the residual term achieved 94.91\%
(Fig.~\ref{fig:main-results}a). As the residual size $g$ grew,
benchmark-calibrated deployment coverage collapsed while response-rank
intervals stayed near nominal coverage (Fig.~\ref{fig:main-results}b). The
conformal result shows response specificity: conformal intervals cover the
response channel on which they are calibrated, while deployment coverage
requires deployment-channel calibration or a residual-response certificate.
Mahalanobis OOD scores and bootstrap uncertainty were likewise nearly
uncorrelated with the completion gap because they measured input novelty or
benchmark-channel variance rather than the benchmark-null deployment direction.

The failure is not merely lower accuracy; it persists at zero benchmark error.
Setting the benchmark-channel error term to zero removes $\delta_c$ from the
interval, but the residual term remains:
\[
|Y_\star(c)-\hat y_c|\leq R_cg .
\]
Thus exact knowledge of the measured benchmark response still leaves completion
cost whenever the deployment direction has a benchmark-null component and the
admissible fiber has positive radius. With $\delta_c=0$, the certified fraction
fell from 100\% at $g=0$ to 69.9\% at $g=0.5$ and 45.4\% at $g=1.0$; completing
the response direction kept certification at 100\%
(Fig.~\ref{fig:main-results}d).

The same geometry produced a leaderboard inversion. Model A had the best
benchmark mean absolute error (MAE) but certified only 6\% of its top-100
candidates; Model E had worse benchmark MAE but certified all top-100
candidates because its response span covered the deployment probe
(Fig.~\ref{fig:main-results}c). Across 50
runs, the best-MAE model was not the best-certifying model in all structured
runs and in 72\% of equal-noise null runs. A benchmark can therefore reward a
model specialized to measured response coordinates while penalizing one that
spans directions needed for deployment.

Completion can also come from trusted structure rather than direct measurement
of the deployment response. In a nonlinear constrained-fiber experiment,
benchmark evidence alone led to false decisions on 15.0\% of candidates; an
ambient response-rank certificate certified 16.1\% with zero false
certificates. Adding the valid constraint $h\approx\sin(\pi b)$, where $b$ is
scalar benchmark evidence and $h$ is an unobserved hidden response coordinate,
refined the admissible set and raised certification to 97.0\% with zero false
certificates.

\subsection*{Completion reduces false decisions and changes model choice}

The consequence is not only residual ambiguity: in a representative JARVIS
split, the benchmark-MAE model made 613 false band-gap decisions and the
response-selected certify-and-acquire workflow made none. More generally, if
benchmark evidence does not determine the deployment action, benchmark actions
can make false decisions. Exact finite audits certify released candidate sets;
operational replays estimate fibers from calibration data and apply them to
held-out candidates, so they measure empirical policy risk rather than
mathematical zero-error certification on an unobserved population.

\begin{figure}[t]
\centering
\includegraphics[width=\linewidth]{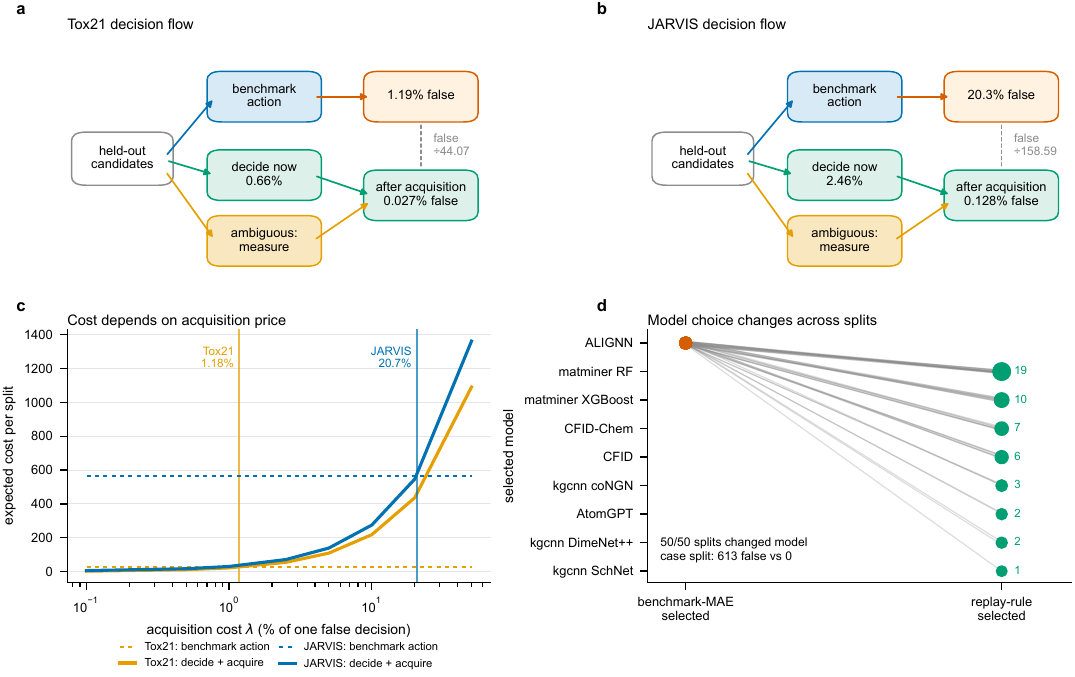}
\caption{\textbf{Completion reduces false decisions and reverses model choice.}
\textbf{a}, In 100 held-out Tox21 splits, the response-certification rule plus
acquisition of ambiguous fibers reduces empirical false SR-p53 decisions from
1.19\% to 0.027\%. \textbf{b}, In JARVIS, 50 held-out splits for each of 15
public models show that the response-certification rule plus acquisition
reduces empirical false band-gap decisions from 20.3\% to 0.128\%.
\textbf{c}, Expected cost as a function of acquisition cost, with break-even
points at 1.18\% for Tox21 and 20.7\% for JARVIS. \textbf{d}, Across JARVIS splits,
formation-energy MAE and replay-certified yield select different models; in a
representative split, the benchmark-MAE model makes 613 false band-gap
decisions whereas the response-selected certify-and-acquire workflow makes
none. Nonzero false decisions arise from the immediately decided
calibration-certified subset; ambiguous candidates are deferred to acquisition
and are not counted as benchmark-supported decisions. Split intervals summarize
replay variability over repeated partitions of the same finite datasets rather
than independent population confidence intervals.}
\label{fig:operational}
\end{figure}

In Tox21, a benchmark-fiber majority policy decided every held-out compound and
made empirical false SR-p53 decisions on 1.19\% of candidates (10th--90th
percentile, 1.00--1.37\%). A conservative response-certification rule with
minimum support 50 decided 0.66\% immediately and sent the rest to SR-p53
acquisition; after acquisition, the empirical false-decision rate was 0.027\%
(0.00--0.18\%). In JARVIS, formation-energy predictions were the benchmark
observable and band-gap energy ($E_{\mathrm{gap}}>1.0\,\mathrm{eV}$) was the
deployment claim. Across 15
public models and 50 held-out splits per model, local benchmark-majority action
made empirical false band-gap decisions on 20.3\% of candidates
(19.2--21.6\%). The response-certification rule immediately decided 2.46\%,
sent the rest to acquisition and reduced empirical false decisions to 0.128\%
(0.00--0.29\%; Fig.~\ref{fig:operational}). Model choice also changed:
benchmark MAE always selected an exact-MAE formation-energy model, whereas
response-certified held-out behaviour selected a different model in all 50
splits.

We converted the replay into a costed deployment decision, measuring costs in
units of one false deployment decision and charging acquisition for each
ambiguous candidate. In JARVIS, certify-then-acquire avoided 562.2 false
band-gap decisions per split on average and remained lower-cost whenever one
band-gap acquisition cost less than 20.7\% of a false deployment decision
(10th--90th percentile, 19.5--21.9\%). Tox21 avoided 25.6 false decisions per
split and broke even below 1.18\% (1.00--1.37\%). An asymmetric loss sweep
weights false positives and false negatives separately and reduces to these
break-even values when the two costs are equal. Ambiguous
candidates are not failures in this workflow; they are acquisition requests
when benchmark evidence does not determine the deployment response.

The next section traces these operational failures to the fiber structure of
reported benchmark evidence across domains.

\subsection*{Released evidence leaves mixed deployment fibers across domains}

Across public audits the pattern was systemic: Tox21 had 97.9\% mixed
seven-assay fibers, and the main 20-quantile Matbench/JARVIS audits had 0\%
median certifiable fraction. The purpose of these audits is not to show that
distinct scientific properties differ. It is to explain why evidence actually
released or scored by a benchmark can fail in the operational replays above. A
mixed fiber, or benchmark twin, is a pair of candidates with the same reported
evidence but different deployment actions. Tox21 has exact binary assay fibers,
spin-defect screening has reported substrate completions, the vision audit uses
finite clean-prediction fibers, and Matbench Discovery and JARVIS require
declared finite-resolution fibers for continuous predictions. The
domain-specific scored evidence maps and fiber rules are specified in
Supplementary Table~1.

Across this range, mixed fibers were common. In the lightweight vision sanity
check, clean prediction, clean correctness and decile-binned clean confidence
left a median 22.7\% of corruption-robustness claims ambiguous across eight
classifiers and 25 splits\cite{pedregosa2011scikit}. In Tox21, seven
nuclear-receptor assays were the benchmark response and SR-p53 the deployment
endpoint\cite{wu2018moleculenet}, producing the 97.9\% mixed-fiber rate above.
In spin-defect
screening, bare-host coherence was the benchmark response and substrate
viability the deployment claim\cite{awschalom2018quantum,seo2017quantum,
toriyama2024screening,toriyama2024zenodo}; 43 of 187 valid hosts (23.0\%)
were response-ambiguous.

We also computed a loss-aware ambiguity score: the Bayes error of the best
fiber-wise benchmark-only action, normalized by the global majority-action
error. This prevalence-normalized residual ambiguity was 0.65 in vision, 0.85
in Tox21 and 0.98 as a conservative spin-defect lower bound, so the mixed
fibers are not merely visual artifacts of label imbalance.

\begin{figure}[t]
\centering
\includegraphics[width=\linewidth]{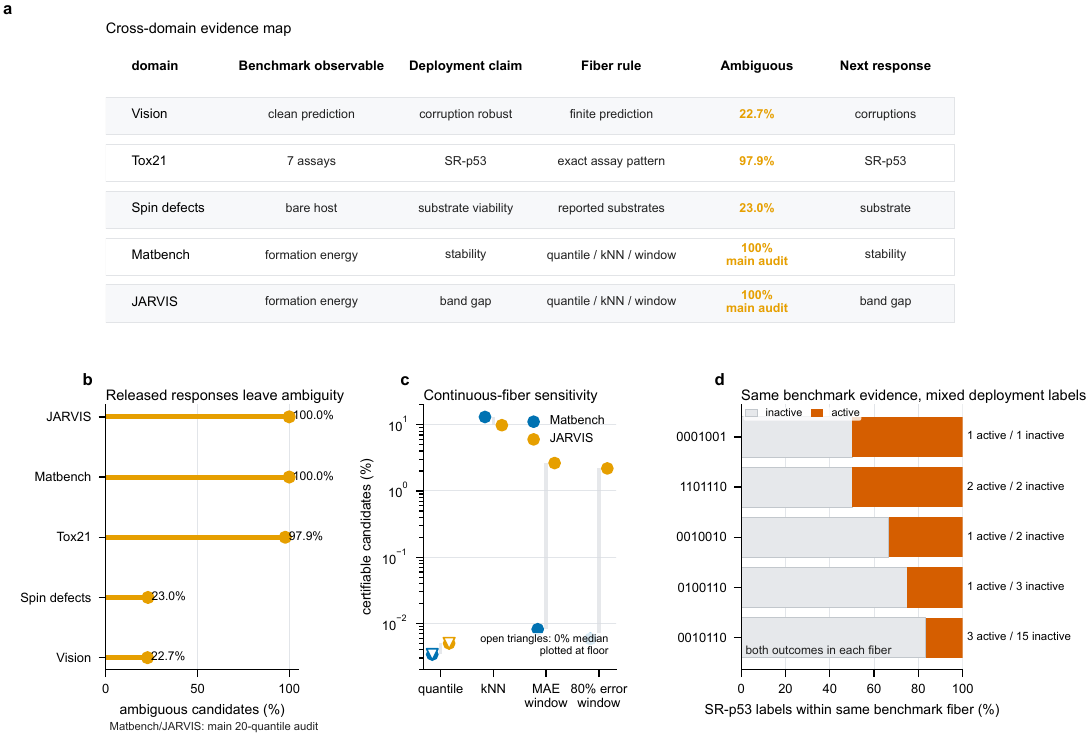}
\caption{\textbf{Released evidence leaves mixed deployment fibers
across domains.} \textbf{a}, Evidence maps linking each benchmark
observable to the deployment claim, fiber rule, ambiguity fraction and next
response. \textbf{b}, Ambiguous fractions across vision, toxicology,
spin-defect screening and materials benchmarks. Matbench and JARVIS use the
main 20-quantile released-prediction audit; panel c summarizes
finite-resolution sensitivity. \textbf{c}, Low certifiable fractions persist
under quantile, nearest-neighbour and model-specific prediction-error windows;
open triangles mark true 0\% medians plotted at the log-scale floor. Full
sensitivity values are reported in Supplementary Table~2. \textbf{d}, Tox21
example showing identical seven-assay benchmark evidence with mixed SR-p53
labels.}
\label{fig:audits}
\end{figure}

Materials-property leaderboards showed the same pattern. Under the main
20-quantile released-prediction audit, the median certifiable fraction was 0\%
for both Matbench Discovery stability labels across 67 public prediction files
and JARVIS band-gap threshold labels across 15 public models
\cite{riebesell2024matbench,wang2021predicting,choudhary2020joint}.
Sensitivity analyses over quantile resolution, nearest-neighbour fibers and
calibration-error windows showed the same qualitative conclusion:
released formation-energy predictions alone left the stability and band-gap
threshold claims largely unresolved at the declared finite resolutions
(Fig.~\ref{fig:audits}c). Under the main 20-quantile rule, normalized residual
ambiguity was 1.00 for Matbench and 0.89 for JARVIS. These audits isolate
released-response evidence, not richer descriptors, domain models or auxiliary
measurements.

\subsection*{Locked completion selects the deployment-relevant probe}

\begin{figure}[t]
\centering
\includegraphics[width=\linewidth]{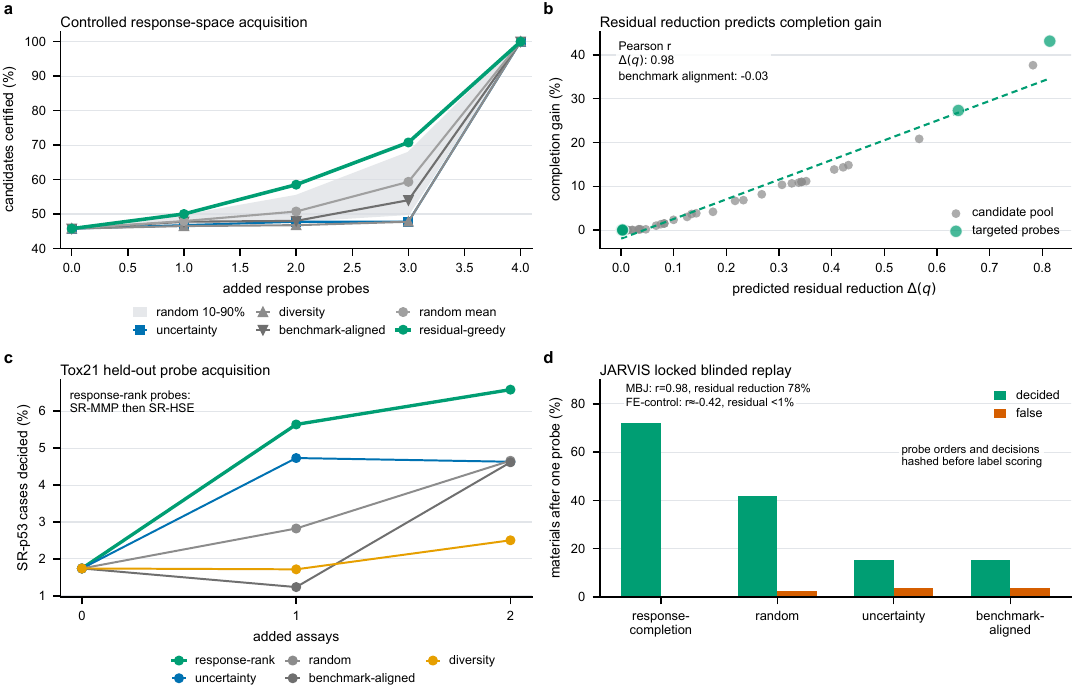}
\caption{\textbf{Locked completion selects the deployment-relevant probe.} \textbf{a},
In the controlled response space,
residual-greedy acquisition certifies more candidates than uncertainty,
diversity, benchmark-aligned or random sampling; points show policy means over
50 seeds. \textbf{b}, Across candidate probes for a single declared deployment
response, the predicted residual reduction $\Delta(q)$ tracks realized
completion gain, while benchmark alignment does not. \textbf{c}, In held-out
Tox21 probe acquisition over 200 random splits, response-rank selects SR-MMP
first and SR-HSE second, deciding more SR-p53 cases under the
calibration-certification rule than baselines after one or two added assays.
\textbf{d}, Locked, label-blinded JARVIS replay over 200 calibration/test
splits. Bars show calibration-certified held-out decisions after one added
probe; false portions are empirical false decisions among all held-out
candidates after the locked rule. Response-completion selects real MBJ band gap
and decides 72.0\% with 0.54\% empirical false decisions; benchmark-aligned and
uncertainty policies select a synthetic formation-energy-like control and
decide 15.1\% with 3.69\% empirical false decisions.}
\label{fig:active}
\end{figure}

In the locked label-blinded JARVIS replay, one completion-selected probe
decided 72.0\% of held-out band-gap cases with 0.54\% empirical false decisions,
versus 15.1\% with 3.69\% for benchmark-aligned or uncertainty policies. It
selected real modified Becke--Johnson (MBJ) band gap in every split, while
benchmark-aligned and uncertainty policies selected the formation-energy-like
control (Fig.~\ref{fig:active}). Completion curves select response information
that makes an action decidable, not merely measurements that are uncertain,
diverse or benchmark-aligned. In the controlled response space, three
residual-greedy probes raised certification from 45.8\% to 70.8\%, outperforming
uncertainty or diversity (47.8\%), benchmark-aligned sampling (54.0\%) and
random sampling (59.3\%; Fig.~\ref{fig:active}). Predicted residual reduction
tracked realized completion gain across probes (Pearson $r=0.976$), whereas
benchmark alignment did not ($r=-0.029$).

In Tox21, response-rank selected SR-MMP then SR-HSE as completion probes for
the held-out SR-p53 action. One and two added assays decided 5.64\% and 6.58\%
of held-out compounds, above uncertainty, random, diversity and benchmark-aligned
baselines; label permutation reduced the gain. A supervised selective baseline
showed how richer evidence legitimately changes the map: with only the seven
benchmark assays it decided 1.55\% of held-out compounds, whereas adding
simplified molecular-input line-entry system (SMILES) descriptors decided 92.0\%
with 0.95\% empirical false decisions.

In a cost-weighted JARVIS companion pool, completion per cost selected a
low-cost residual probe in 92.0\% of splits, while benchmark-aligned acquisition
selected the cheap formation-energy-like probe.

\FloatBarrier
\section*{Discussion}

Deployment-complete benchmarking changes the unit of evaluation from a score to
a score--claim pair. A benchmark score becomes meaningful for deployment only
after specifying the action it is meant to support and the evidence through
which the benchmark observes candidates. The factorization condition gives an
exact diagnostic for this relationship; completion curves quantify the cost of
repairing it when it fails. The output is an auditable evidence package: score,
evidence map, supported action, ambiguous fraction, completion curve and next
measurement.

Accuracy, calibration, conformal prediction, uncertainty estimation,
out-of-distribution detection and active learning remain valuable, but they
answer different questions: ranking, reliability, novelty or label efficiency
for a measured response. Completion asks whether the evidence package
determines the action. This connects benchmark reporting to sufficiency and
experiment comparison\cite{fisher1922mathematical,blackwell1953equivalent},
partial identification\cite{manski1990nonparametric,manski2003partial,
tamer2010partial}, value-of-information and optimal design
\cite{lindley1956information,howard1966information,chaloner1995bayesian,
pukelsheim2006optimal}, as well as property elicitation, robust decision-making
and selective classification\cite{gneiting2011making,lambert2008eliciting,
frongillo2015vector,chow1970optimum,geifman2017selective,vovk2005algorithmic,
lei2018distribution,tibshirani2019conformal,hendrycks2017baseline,
ovadia2019uncertainty,koh2021wilds,gulrajani2021search}.

The experiments show why this standard matters. In controlled response spaces,
zero benchmark error and valid benchmark-channel conformal coverage do not
imply deployment-channel completeness. Across modalities, reported benchmark
evidence leaves mixed fibers and residual deployment risk. In Tox21 and JARVIS
replays, completion-aware policies select different measurements, reduce false
decisions, defer ambiguous cases to acquisition and can change model choice.
The framework is evidence-relative by design: descriptors, model internals,
auxiliary labels, physical constraints or prospective measurements change the
evidence map and can lower completion cost. For benchmarks used in deployment,
procurement or scientific screening, reports should state the score, evidence
map, supported action, ambiguous fraction, completion curve and remaining
acquisition cost.

\clearpage
\section*{Methods}

\medskip\noindent\textbf{Certification terminology.}
We use three certification terms. Exact certification refers to finite
released-set fibers whose deployment labels are known and pure; this is a
deterministic statement about the declared candidate set. Calibration
certification refers to a rule learned from calibration fibers and applied to
held-out candidates; it produces an empirical false-decision rate and is not a
formal zero-error population guarantee. Predicted certification refers to
interval or residual-radius rules derived from declared modeling assumptions.
All held-out replay results are calibration-certified policy evaluations
unless explicitly labelled exact finite certifications.

\medskip\noindent\textbf{Claim-completeness and finite fibers.}
Let $\mathcal{S}$ be an admissible system class, let
$E:\mathcal{S}\to\mathcal{Z}$ be the declared benchmark evidence map and let
$D:\mathcal{S}\to\mathcal{A}$ be the deployment action. The evidence-compatible
fiber at $z$ is $\mathcal{C}_z=\{s\in\mathcal{S}:E(s)=z\}$. The benchmark is
claim-complete for $D$ on a fiber when $D$ is constant on $\mathcal{C}_z$.
The factorization theorem follows from the standard quotient argument. If
$D=\phi\circ E$, then $E(s_0)=E(s_1)$ implies
$D(s_0)=\phi(E(s_0))=\phi(E(s_1))=D(s_1)$, so $D$ is constant on benchmark
fibers. Conversely, if $D$ is constant on each nonempty fiber, choose any
$s_z\in E^{-1}(z)$ and define $\phi(z)=D(s_z)$. The definition is
well-defined because the fiber is pure, and $\phi(E(s))=D(s)$ for every
$s\in\mathcal{S}$. The indistinguishability statement is the same condition
written contrapositively. Score-invariance is immediate: if a statistic
computed only from the declared benchmark evidence has the form $T=h(E)$ and
$D=\psi\circ T$, then
$D=\psi\circ h\circ E$, so $D$ would factor through $E$.
For cost-sensitive actions with state-dependent deployment loss $L(a,s)$, this
can be relaxed to an $\epsilon$-robust fiber
condition
\[
\inf_{a\in\mathcal{A}}\sup_{s\in\mathcal{C}_z}
\left\{L(a,s)-\inf_{a'\in\mathcal{A}}L(a',s)\right\}
\leq \epsilon .
\]
In finite audits with candidates $i=1,\ldots,N$, the exact certifiable fraction
is
\[
\Cert(E,D)=
\frac{1}{N}\sum_z |\mathcal{C}_z|\,
\mathbf{1}\{D\text{ is constant on }\mathcal{C}_z\},
\qquad
\Amb(E,D)=1-\Cert(E,D).
\]
For binary deployment actions, we also report the fiber Bayes error of the best
benchmark-only decision rule,
\[
\operatorname{Err}(E,D)=
\frac{1}{N}\sum_z |\mathcal{C}_z|\min\{p_z,1-p_z\},
\qquad
p_z=\Pr(D=1\mid E=z),
\]
and the prevalence-normalized residual ambiguity
\[
\rho(E,D)=
\frac{\operatorname{Err}(E,D)}
{\min\{\Pr(D=1),\Pr(D=0)\}}.
\]
Here $\rho=0$ means the declared evidence completes the binary action, whereas
$\rho=1$ means the best evidence-fiber action has the same error as the global
majority rule for a proper partition. For $k$-nearest-neighbour (kNN) and
error-window sensitivity, neighbourhoods overlap rather than partition the
candidate set, so the reported
decision-risk column is a local-neighbourhood majority error and normalized
values can slightly exceed one. The spin-defect value is reported as a
conservative lower-bound risk because the public file records substrate
completions rather than a full candidate-by-substrate binary table.
For a scalar deployment response $Y_\star:\mathcal{S}\to\mathbb{R}$ and
threshold $\tau$, define
$D_\tau(s)=\mathbf{1}\{Y_\star(s)>\tau\}$. A fiber is certified positive when
$\inf_{s\in\mathcal{C}_z}Y_\star(s)>\tau$, certified negative when
$\sup_{s\in\mathcal{C}_z}Y_\star(s)\leq\tau$ and ambiguous when the threshold
cuts through the fiber. A mixed fiber directly witnesses non-factorization.

\medskip\noindent\textbf{Completion cost.}
Let $\mathcal{Q}$ be a set of possible response probes and let
$E_U(s)=(E(s),\{q(s):q\in U\})$ be the evidence after adding
$U\subseteq\mathcal{Q}$. With additive probe costs $\cost(U)=\sum_{q\in U}c_q$,
the ideal completion curve is
\[
\Gamma_E(b;D)=
\max_{U\subseteq\mathcal{Q}:\,\cost(U)\leq b}\Cert(E_U,D),
\]
and the $\epsilon$-completion cost is
\[
\kappa_\epsilon(E,D)=
\inf\{b:\Gamma_E(b;D)\geq 1-\epsilon\}.
\]
Exact optimization can be combinatorial; the experiments therefore report
completion curves under specified policies and compare them with oracle,
uncertainty, diversity, benchmark-aligned and random baselines.

The loss-aware analogue uses action-label loss $\ell$ and replaces pure-fiber
counting with Bayes risk. For a population distribution on $\mathcal{S}$,
define
\[
R(E,D;\ell)=\inf_{\phi:\mathcal{Z}\to\mathcal{A}}
\mathbb{E}\{\ell(\phi(E(S)),D(S))\}.
\]
For augmented evidence $E_U$, the expected value of evidence is
\[
\Delta_\ell(U)=R(E,D;\ell)-R(E_U,D;\ell),
\]
and a cost-aware completion objective is
\[
V_\ell(U)=R(E_U,D;\ell)+\cost(U).
\]
The one-step expected value of information for a probe $q$ after already
measuring $U$ is
\[
\operatorname{EVI}(q\mid U)=
R(E_U,D;\ell)-R(E_{U\cup\{q\}},D;\ell).
\]
Exact claim-completeness is the zero-risk special case
$R(E,D;\ell_{0/1})=0$ under 0--1 deployment loss.

\medskip\noindent\textbf{Linear response-rank specialization.}
Let $\mathcal{H}$ be a Hilbert space of response functions or latent response
states. Benchmark probes $k_1,\ldots,k_m\in\mathcal{H}$ define evidence
$E_B(s)=(\langle k_1,s\rangle,\ldots,\langle k_m,s\rangle)$ and span
$B=\spanop\{k_1,\ldots,k_m\}$, with orthogonal projector $\PB$. A deployment
response is $Y_\star(s)=\langle\kstar,s\rangle$. If $\kstar\in B$, then
$Y_\star$ is a linear function of benchmark evidence. If $\kstar\notin B$, let
\[
\rstar=(I-\PB)\kstar,\qquad g=\|\rstar\|.
\]
Then $\rstar\neq 0$, $\langle k_j,\rstar\rangle=0$ for each benchmark probe
and, for any scalar $t\neq0$, the two worlds $s_0=0$ and $s_1=t\rstar$ have
identical benchmark evidence but different deployment response:
\[
Y_\star(s_1)-Y_\star(s_0)=t\|\rstar\|^2\neq 0.
\]
Thus $g$ is the norm of the deployment direction invisible to the benchmark.
A nonzero residual is an algebraic invisibility diagnostic in the ambient
response space. Claim-incompleteness on a particular admissible set
$\mathcal{S}$ additionally requires either an explicit pair
$s_0,s_1\in\mathcal{S}$ with equal benchmark evidence and different deployment
action, or a declared feasible-fiber radius that permits variation in the
residual direction. When physical, causal or policy constraints shrink feasible
fibers, the relevant quantity is the feasible deployment-response diameter, not
the ambient residual norm alone.

\medskip\noindent\textbf{Feasible-fiber and interval bounds.}
Let $\mathcal{S}\subset\mathcal{H}$ be an admissible system class and
let $b=(b_1,\ldots,b_m)$ be a benchmark evidence vector. The set
$\mathcal{C}_b=\{s\in\mathcal{S}:\langle k_j,s\rangle=b_j,\ j=1,\ldots,m\}$
is the corresponding benchmark fiber. For any $s,s'\in\mathcal{C}_b$,
\[
|\langle \kstar,s'-s\rangle|\leq g\,\|s'-s\|,
\]
because $s'-s$ is benchmark-null and Cauchy--Schwarz gives
$|\langle \rstar,s'-s\rangle|\leq g\|s'-s\|$. If the admissible class contains
the radius-$R$ step $s+R\rstar/\|\rstar\|$, benchmark measurements are
unchanged and the deployment response changes by $Rg$. Physical constraints may shrink the feasible fiber,
in which case the exact ambiguity is the feasible-fiber diameter rather than
the ambient Hilbert-ball diameter.

If benchmark-channel prediction error is bounded by $\delta_c$ and candidate
$c$ has admissible benchmark-null radius $R_c$, every compatible perturbation
$u\perp B$ with $\|u\|\leq R_c$ satisfies
\[
|\langle\kstar,u\rangle|=
|\langle\rstar,u\rangle|
\leq R_cg,
\]
so
\[
|Y_\star(c)-\hat y_c|\leq \delta_c+R_cg.
\]
Here $\hat y_c$ is the benchmark-based center, or predicted deployment response,
for candidate $c$.
The diagnostic forms
$I_c=[\hat y_c-(\delta_c+R_cg),\hat y_c+(\delta_c+R_cg)]$ and certifies a
threshold action only when $I_c$ lies wholly on one side of the threshold.
$R_c$ is declared before certification from replicate measurement variation,
simulation/model discrepancy, validation fibers or domain tolerance.

\medskip\noindent\textbf{One-step response completion.}
For a candidate response probe $q$, let $q_\perp=(I-\PB)q$. If
$q_\perp=0$, the probe lies inside the existing benchmark span and cannot
reduce the residual. Otherwise the updated residual after adding $q$ is
\[
\rstar(q)=
\rstar-\frac{\langle\rstar,q_\perp\rangle}{\|q_\perp\|^2}q_\perp,
\]
so
\[
g(q)^2=\|\rstar(q)\|^2
=g^2-\frac{\langle\rstar,q_\perp\rangle^2}{\|q_\perp\|^2}.
\]
The squared residual reduction is therefore
\[
\Delta(q)=
\frac{\langle\rstar,q_\perp\rangle^2}{\|q_\perp\|^2},
\]
and the cost-normalized greedy completion probe is
\[
q^\star=\operatorname*{arg\,max}_{q\in\mathcal{Q}}
\frac{\langle\rstar,q_\perp\rangle^2}{\|q_\perp\|^2c_q}.
\]
The selected probe is appended to the benchmark set, intervals are recomputed
and candidate classes updated. Radius sensitivity multiplied all $R_c$ by
0.5, 0.75, 1.0, 1.25, 1.5 and 2.0.

\medskip\noindent\textbf{Local nonlinear extension.}
For differentiable response models $y_k(x)=\langle k,f_\theta(x)\rangle$, let
$J_B(x)$ be the matrix of benchmark gradients and $P_{J_B}$ the projector onto
its row span. Define
$g_J=\|(I-P_{J_B})\nabla y_\star(x)\|$. If
$\|\nabla^2 y_\star\|_2\leq L_\star$ in $\|\Delta x\|\leq R_c$, then every
local benchmark-null direction $J_B(x)\Delta x=0$ satisfies
\[
|y_\star(x+\Delta x)-y_\star(x)|\leq R_cg_J+\tfrac12 L_\star R_c^2.
\]
For benchmark channel $j$ with Hessian bound
$\|\nabla^2 y_j\|_2\leq L_j$ on the same neighbourhood, tangent-null leakage is
bounded by $\tfrac12L_jR_c^2$. The nonlinear ablation used NumPy tanh networks with input
dimension 12, output dimension 8, hidden width 32, depths 1, 2, 4 and 6, and
50 random initializations per depth. Gradients were computed exactly by
backpropagation and local Hessian envelopes were estimated from random
directions. Outputs are
\texttt{nonlinear\_linearization\_ablation.csv} and its summary file.

\medskip\noindent\textbf{Controlled experiments.}
The response-channel transfer experiment used an 8-dimensional response space,
four benchmark probes of effective rank 3, label noise $\sigma=0.05$ and split
conformal prediction with $\alpha=0.05$. OOD detection used Mahalanobis
distance and uncertainty used a 50-member bootstrap ensemble. Repeated
summaries used 50 seeds. The leaderboard experiment used a 10-dimensional
response space, deployment probe
$\kstar=(e_0+\cdots+e_7)/\sqrt{8}$ and five model spans of rank 3, 4, 6, 7 and
8. Structured and equal-noise repetitions are written to
\texttt{revision\_leaderboard\_repeated\_summary.csv}.
The decision-sufficiency generalization experiment used 100 seeds with 1,000
candidates per seed. The benchmark evidence was $b\in[-1,1]$, the deployment
response was $y_\star=b+0.9h$ with threshold $\tau=0.4$, and the admissible
hidden coordinate satisfied $h=\sin(\pi b)+u$, $u\in[-0.12,0.12]$. The ambient
response-rank certificate used only $h\in[-1.2,1.2]$. Outputs are cached in the
\texttt{outputs/} directory with the prefix \texttt{decision\_sufficiency}.
The zero-benchmark-error control fixed the same certification rule while
setting $\delta_c=0$ and sweeping $g\in[0,1]$; the residual-reduction control
evaluated each candidate probe against one declared deployment response and
compared the theoretical $\Delta(q)$ with realized certified-fraction gain.
Outputs are
\texttt{zero\_benchmark\_error\_control.csv},
\texttt{residual\_reduction\_completion\_gain.csv} and
\texttt{residual\_reduction\_completion\_gain\_summary.csv}.

\medskip\noindent\textbf{Operational replays.}
The replay script used held-out calibration/deployment splits and should be
read as an empirical risk evaluation of a completion-aware policy. In Tox21,
100 random splits used 50\% calibration compounds to define seven-assay fibers
and their calibration SR-p53 labels to estimate fiber purity; 50\% held-out
compounds were then used to evaluate SR-p53 decisions. A held-out fiber could
certify only if it had at least 50 calibration compounds and a single
calibration SR-p53 label. More generally, for a calibration fiber $z$ with
$n_z$ examples and empirical disagreement rate $\hat p_z$, a
calibration-certified rule can require a one-sided upper confidence bound
$p_z^+$ to satisfy $p_z^+\leq\tau$. With zero observed disagreements,
the Clopper--Pearson bound is $p_z^+=1-\delta_z^{1/n_z}$; allocating
$\delta_z=\delta/m$ across $m$ tested fibers gives simultaneous conditional
error control under exchangeability within fibers. The operational replays use
the simpler unanimous-support rule and report held-out empirical risk; the
confidence-bound version is the corresponding finite-sample calibration
certificate. Under a Bernoulli interpretation without multiplicity correction,
50 unanimous labels exclude discordance rates above $1-0.05^{1/50}=5.8\%$ at
95\% confidence. In
JARVIS, 50 splits per model used formation-energy prediction windows with
half-width equal to calibration MAE; a held-out material was certified only
when all calibration materials in its window shared the same band-gap threshold
label.

The costed replay paired each benchmark-action row with the corresponding
certify-then-acquire row. Benchmark action incurred one unit for every false
deployment decision. Certify-then-acquire incurred the same false-decision cost
plus $\lambda$ units for every ambiguous candidate sent to acquisition. The
reported break-even acquisition cost is
$(E_\mathrm{benchmark}-E_\mathrm{certify+acquire})/A$, where $E$ is the number
of false decisions and $A$ is the number of acquired ambiguous candidates.
This symmetric convention is the special case of an asymmetric deployment-loss
calculation. If $C_{\rm FP}$ and $C_{\rm FN}$ are false-positive and
false-negative costs and $C_{\rm acq}$ is the cost of acquiring the missing
response for one deferred candidate, then
\[
L_{\rm bench}
= C_{\rm FP}\,\mathrm{FP}_{\rm bench}
+ C_{\rm FN}\,\mathrm{FN}_{\rm bench},
\]
whereas
\[
L_{\rm comp}
= C_{\rm FP}\,\mathrm{FP}_{\rm comp}
+ C_{\rm FN}\,\mathrm{FN}_{\rm comp}
+ C_{\rm acq}N_{\rm defer}.
\]
Completion is lower-cost whenever
\[
C_{\rm acq}
<
\frac{
C_{\rm FP}(\mathrm{FP}_{\rm bench}-\mathrm{FP}_{\rm comp})
+ C_{\rm FN}(\mathrm{FN}_{\rm bench}-\mathrm{FN}_{\rm comp})
}{N_{\rm defer}}.
\]
Figure~\ref{fig:operational} reports the symmetric case
$C_{\rm FP}=C_{\rm FN}=1$.
The asymmetric sweep over $C_{\rm FP}/C_{\rm FN}\in[0.1,10]$ is written to
\path{asymmetric_cost_break_even.csv} and plotted in
\path{asymmetric_cost_heatmap.pdf}.
Case studies were chosen from held-out splits by the largest difference
between the certified error of the benchmark-MAE model and the
certification-selected model.

The Tox21 response-probe acquisition campaign used the same complete-label
compounds, the same SR-p53 deployment endpoint and the same minimum support of
50 calibration compounds. The initial response panel was the seven
nuclear-receptor assays; the candidate probes were SR-ARE, SR-ATAD5, SR-HSE and
SR-MMP. For each of 200 random splits, policies chose probes using calibration
compounds and calibration SR-p53 labels only, then scored held-out compounds
after the probe order was fixed. The response-rank policy chose the assay that
maximized the supported pure SR-p53 fiber fraction on calibration data; the
oracle chose the assay that maximized held-out certification and is reported
only as an upper bound. The permutation control used the same splits but
permuted calibration SR-p53 labels for probe selection only; after the probe
order was frozen, certificate maps were built from real calibration labels and
evaluated on held-out compounds. Outputs are
\texttt{tox21\_probe\_permutation\_control.csv} and
\texttt{tox21\_probe\_permutation\_control\_summary.csv}.

The Tox21 deployment-label selective baseline used the same 100 calibration/test
splits and the same held-out SR-p53 endpoint. We trained logistic-regression
classifiers on calibration SR-p53 labels with two evidence maps: the seven
nuclear-receptor assays alone, and those assays plus SMILES character
$n$-gram descriptors (2--4 grams, capped at 2,048 features). A held-out
compound was decided when the predicted class probability exceeded a declared
confidence threshold; otherwise it was deferred. Thresholds
$0.5,0.6,0.7,0.8,0.9$ and $0.95$ were evaluated. Outputs are
\path{tox21_selective_baseline.csv} and
\path{tox21_selective_baseline_summary.csv}. This baseline is an evidence
refinement using deployment labels and richer descriptors, not a statistic of
the original seven-assay evidence map.

The JARVIS label-blinded probe-completion experiment used the 187 materials
present in all three JARVIS-Leaderboard reference files for formation energy
per atom, optB88vdW (optimized Becke88 van der Waals) band gap and MBJ
hybrid-functional band gap. The deployment label was band-gap energy
($E_{\mathrm{gap}} > 1.0\,\mathrm{eV}$; 33.7\% viable). Candidate probes were a
real JARVIS MBJ band-gap response (real JARVIS density-functional-theory (DFT)
calculation, $r = 0.98$ with the deployment target) and a synthetic
formation-energy-correlated control response (formation energy plus Gaussian
noise with standard deviation $0.3\sigma_\mathrm{fe}$, where
$\sigma_\mathrm{fe}$ is the formation-energy standard deviation, mimicking a
benchmark-aligned measurement). In each of 200 calibration/test
splits (70/30), the policy chose the probe from calibration materials only. The
chosen probe, held-out material decisions and campaign manifest were then
written to JSON and hashed before held-out optB88vdW labels were scored.
Certification used 2-D quantile fibers (8 bins per dimension, minimum fiber
size 3). This replay is a response-channel validation experiment rather than a
cost claim about MBJ as the universally appropriate measurement; in a real
screening workflow, the completion probe depends on simulation or measurement
cost. Results are written to
\texttt{outputs/jarvis\_blinded\_completion\_summary.csv}; the locked JSON
artifacts and their hashes use the \texttt{locked\_jarvis\_blind\_*} prefix.

The cost-weighted JARVIS companion experiment used the same 187-material
intersection and deployment threshold, but expanded the candidate pool with
synthetic measured response channels whose cost, benchmark alignment,
deployment-residual alignment and noise were controlled. The pool contained a
cheap benchmark-aligned probe, a mixed low-cost probe, a low-cost residual
probe, a noisy band-gap proxy, an expensive precise residual probe and real MBJ
band gap. The completion-per-cost policy maximized the calibration
response-rank score divided by declared probe cost; response-rank ignored cost;
benchmark-aligned selected maximum formation-energy alignment; uncertainty
selected maximum calibration variance; random sampled uniformly; and the
oracle-per-cost policy maximized held-out certification gain per cost and is
reported only as an upper bound. Outputs are
\path{outputs/jarvis_cost_weighted_probe_pool_summary.csv} and
\path{outputs/jarvis_cost_weighted_probe_pool.pdf}.

\medskip\noindent\textbf{Public finite-fiber audits.}
The vision audit used the scikit-learn handwritten-digits dataset and eight
classifiers: logistic regression, linear SVM, RBF SVM, random forest,
extra-trees, $k$-nearest neighbours, a one-hidden-layer multilayer perceptron
and Gaussian naive Bayes\cite{pedregosa2011scikit}. Across 25 stratified
splits, the benchmark response was clean-image correctness. The deployment
response was correctness on the clean image and four deterministic corruptions:
Gaussian noise, Gaussian blur, central occlusion and subpixel shift. Finite
benchmark fibers were defined by clean predicted class, clean correctness and
decile-binned clean confidence. The Tox21 audit downloaded the MoleculeNet CSV
from the public DeepChem host,
kept compounds with complete labels for seven nuclear-receptor assays and
SR-p53, grouped exact assay patterns and classified fibers by unanimity of the
held-out endpoint\cite{wu2018moleculenet}. The spin-defect audit downloaded
Toriyama et al. Zenodo data, merged bare-host and heterostructure $T_2$ records,
and classified reported substrate completions under a $T_2>1$ ms viability
threshold\cite{toriyama2024zenodo}. The Matbench audit downloaded public
Figshare prediction files and the WBM test-set summary, paired formation-energy
predictions with stability labels, and evaluated quantile, nearest-neighbour
and error-window fibers\cite{riebesell2024matbench,wang2021predicting}. The
JARVIS audit downloaded public prediction files for formation energy and
optB88vdW band gap from the JARVIS-Leaderboard GitHub repository and evaluated
an $E_{\mathrm{gap}}>1.0\,\mathrm{eV}$ threshold\cite{choudhary2020joint}. Continuous
audits were reported only after declaring finite resolutions. For Matbench and
JARVIS, quantile fibers used 5, 10, 20, 40, 80 and 100 bins;
nearest-neighbour fibers used $k=10,25,50,100$; and error-window fibers used
each model's mean absolute error, 80th-percentile absolute error and
95th-percentile absolute error as tolerances.
Supplementary Table~2 reports certifiable fractions, fiber sizes,
decision-risk error and normalized residual ambiguity for these declared
finite-resolution rules. These tolerance choices audit released predictions;
structure and descriptors can be added as richer evidence maps.

\FloatBarrier

\clearpage
\section*{Supplementary information}
\begingroup
\singlespacing

\begin{table}[htbp]
\centering
\scriptsize
\setlength{\tabcolsep}{2pt}
\begin{tabular}{L{0.09\linewidth}L{0.13\linewidth}L{0.19\linewidth}L{0.20\linewidth}L{0.16\linewidth}L{0.10\linewidth}}
\toprule
Domain & Unit $s$ & Evidence map $E(s)$ & Evidence added only in completion/richer analyses &
Deployment action $D(s)$ & Fiber rule \\
\midrule
Vision &
Image--classifier instance &
Clean predicted class, clean correctness and clean-confidence decile &
Corrupted-image responses &
Corruption-robust correctness &
Exact finite fibers \\
Tox21 &
Compound &
Seven nuclear-receptor assay labels &
SR assays, molecular structure and fingerprints &
SR-p53 label &
Exact assay pattern \\
Spin defects &
Host/substrate candidate &
Bare-host response &
Substrate-specific response &
Substrate viability &
Reported substrate completions \\
Matbench &
Material--model prediction row &
Released formation-energy prediction &
Structure, composition, model internals and stability label &
Stability decision &
Finite-resolution fibers \\
JARVIS &
Material--model prediction row &
Released formation-energy prediction &
Band-gap probes before completion, structure, composition and model internals &
$E_{\rm gap}>1.0$ eV &
Finite-resolution fibers \\
\bottomrule
\end{tabular}
\caption*{\textbf{Supplementary Table 1. Evidence maps used in the public
audits.} Each row specifies the scored evidence map, the deployment action and
the fiber rule. Information outside the scored evidence map can be added as
richer evidence or response-completion probes, changing the completion cost.}
\end{table}

\begin{table}[p]
\centering
\scriptsize
\setlength{\tabcolsep}{2.5pt}
\resizebox{\linewidth}{!}{%
\begin{tabular}{lllrrrrr}
\toprule
Domain & Fiber rule & Parameter & Cert. \% & IQR \% & Median fiber & Decision-risk \% & $\rho$ \\
\midrule
Matbench & Quantile & 5 & 0.00 & 0.00--0.00 & 51,391 & 18.79 & 1.00 \\
Matbench & Quantile & 10 & 0.00 & 0.00--0.00 & 25,696 & 18.79 & 1.00 \\
Matbench & Quantile & 20 & 0.00 & 0.00--0.00 & 12,848 & 18.79 & 1.00 \\
Matbench & Quantile & 40 & 0.00 & 0.00--0.00 & 6,424 & 18.79 & 1.00 \\
Matbench & Quantile & 80 & 0.00 & 0.00--0.00 & 3,212 & 18.79 & 1.00 \\
Matbench & Quantile & 100 & 0.00 & 0.00--0.00 & 2,570 & 18.79 & 1.00 \\
Matbench & kNN & 10 & 25.43 & 23.95--26.74 & 10 & 18.04 & 0.96 \\
Matbench & kNN & 25 & 13.09 & 11.43--15.24 & 25 & 18.58 & 0.99 \\
Matbench & kNN & 50 & 8.58 & 7.09--11.58 & 50 & 18.71 & 1.00 \\
Matbench & kNN & 100 & 5.18 & 4.38--9.49 & 100 & 18.75 & 1.00 \\
Matbench & Error window & MAE & $<0.01$ & $<0.01$--0.01 & 196,278 & 18.95 & 1.01 \\
Matbench & Error window & 80\% abs. error & $<0.01$ & $<0.01$--$<0.01$ & 217,152 & 19.09 & 1.02 \\
Matbench & Error window & 95\% abs. error & $<0.01$ & $<0.01$--$<0.01$ & 252,058 & 18.82 & 1.00 \\
\midrule
JARVIS & Quantile & 5 & 0.00 & 0.00--0.00 & 1,114 & 21.55 & 1.00 \\
JARVIS & Quantile & 10 & 0.00 & 0.00--0.00 & 557 & 19.40 & 0.90 \\
JARVIS & Quantile & 20 & 0.00 & 0.00--0.00 & 279 & 19.27 & 0.89 \\
JARVIS & Quantile & 40 & 0.00 & 0.00--0.00 & 139 & 19.10 & 0.89 \\
JARVIS & Quantile & 80 & 1.26 & 1.25--2.50 & 70 & 18.93 & 0.88 \\
JARVIS & Quantile & 100 & 2.01 & 1.50--3.01 & 56 & 18.95 & 0.88 \\
JARVIS & kNN & 10 & 27.08 & 26.59--27.26 & 10 & 17.68 & 0.82 \\
JARVIS & kNN & 25 & 9.83 & 8.78--10.97 & 25 & 18.57 & 0.86 \\
JARVIS & kNN & 50 & 2.33 & 2.00--3.85 & 50 & 18.90 & 0.88 \\
JARVIS & kNN & 100 & 0.00 & 0.00--0.54 & 100 & 19.04 & 0.88 \\
JARVIS & Error window & MAE & 2.64 & 1.32--3.89 & 128 & 18.89 & 0.88 \\
JARVIS & Error window & 80\% abs. error & 2.19 & 1.03--3.01 & 159 & 18.96 & 0.88 \\
JARVIS & Error window & 95\% abs. error & 0.74 & 0.55--1.35 & 501 & 19.12 & 0.89 \\
\bottomrule
\end{tabular}}
\caption*{\textbf{Supplementary Table 2. Continuous-fiber decision-risk
sensitivity.} Values are medians across 67 Matbench Discovery prediction files
or 15 JARVIS public models. IQR denotes the 25th--75th percentile interval
across files or models. For quantile fibers, decision-risk error is the Bayes
error of the best partition-wise action under the declared evidence rule. For
kNN and error-window neighbourhoods, it is local-neighbourhood majority error;
because these neighbourhoods overlap and do not define a global partition, the
normalized value can slightly exceed one when the local evidence is less
informative than the global majority rule. The source table is
\texttt{outputs/continuous\_fiber\_decision\_risk\_summary.csv}.}
\end{table}

\FloatBarrier
\endgroup

\section*{Data availability}
Public datasets are downloaded or loaded at runtime by the scripts:
scikit-learn handwritten digits, MoleculeNet Tox21, Toriyama et al. Zenodo
spin-defect data, Matbench Discovery Figshare/WBM test-set files and JARVIS-Leaderboard
GitHub files. Numerical summaries cited in the manuscript are cached in
\texttt{outputs/}. The label-blinded JARVIS replay additionally writes locked
probe-order, decision and manifest JSON files with SHA-256 hashes before
held-out deployment labels are scored.

\section*{Code availability}
The reusable BenchCert tool for deployment-completeness audits is available at
\url{https://github.com/E-zClap/benchcert}. Code to reproduce the analyses,
figures and tables in this manuscript is available at
\url{https://github.com/E-zClap/benchcert-reproducibility}. The
reproducibility repository includes installation instructions, dependency
specifications, scripts for all experiments and cached numerical outputs. The
locked JARVIS replay is generated by
\texttt{scripts/jarvis\_blinded\_completion\_campaign.py}. The command
\texttt{bash reproduce\_all.sh} regenerates the cached outputs and manuscript
figures from public data, including the continuous-fiber decision-risk table,
prevalence-normalized ambiguity summary, Tox21 selective baseline and
asymmetric cost sweep. The cost-weighted JARVIS companion pool is generated by
\texttt{scripts/jarvis\_cost\_weighted\_probe\_pool.py}.

\section*{Acknowledgements}
This research was supported by JSPS KAKENHI Grant Number 24K21730.


\section*{Competing interests}
The authors declare no competing interests.

\bibliographystyle{unsrtnat}
\bibliography{references}

\end{document}